\documentclass{article}

\PassOptionsToPackage{numbers, compress}{natbib}

\usepackage{amsmath}
\usepackage[final]{neurips_2025_ml4ps}

\usepackage[utf8]{inputenc} 
\usepackage[T1]{fontenc}    
\usepackage{hyperref}       
\usepackage{url}            
\usepackage{booktabs}       
\usepackage{amsfonts}       
\usepackage{nicefrac}       
\usepackage{microtype}      
\usepackage{xcolor}         
\usepackage{graphicx}       
\usepackage{amsmath}        
\usepackage{enumitem}       
\usepackage{csquotes}       
\usepackage{tabularx}
\usepackage{booktabs}

\title{Generalizing PDE Emulation with Equation-Aware Neural Operators}

\author{%
  Qian-Ze Zhu\\
  School of Engineering and Applied Sciences\\
  Harvard University\\
  Cambridge, MA, USA\\
  Google Research\\
  Cambridge, MA, USA\\
  \texttt{qianzezhu@g.harvard.edu} \\
  \And
  Paul Raccuglia \\
  Google Research\\
  Cambridge, MA, USA\\
  \texttt{praccu@google.com} \\
  \AND
  Michael P. Brenner\\
  School of Engineering and Applied Sciences\\
  Harvard University\\
  Cambridge, MA, USA\\
  Google Research\\
  Cambridge, MA, USA\\
  \texttt{brenner@seas.harvard.edu} \\
}

\begin{document}

\maketitle

\begin{abstract}
    Solving partial differential equations (PDEs) can be prohibitively expensive using traditional numerical methods. Deep learning-based surrogate models typically specialize in a single PDE with fixed parameters. 
    We present a framework for equation-aware emulation that generalizes to unseen PDEs, conditioning a neural model on a vector encoding representing the terms in a PDE  and their coefficients. 
    We present a baseline of four distinct modeling technqiues, trained on a family of 1D PDEs from the APEBench suite. 
    Our approach achieves strong performance on parameter sets held out from the training distribution, with strong stability for rollout beyond the training window, and generalization to an entirely unseen PDE.
    This work was developed as part of a broader effort exploring AI systems that automate the creation of expert-level empirical software for scorable scientific tasks \cite{aygun2025ai}.
    The codebase and data are available in \href{https://github.com/google-research/generalized-pde-emulator}{\texttt{https://github.com/google-research/generalized-pde-emulator}}.
\end{abstract}

\section{Introduction}

Partial differential equations (PDEs) are the mathematical foundation for modeling a vast range of physical phenomena. Traditional numerical methods, while accurate, are often computational expensive, scaling poorly for high-resolution or long-term simulations. This motivated the development of deep learning-based surrogate models, which learn to approximate the solution operator of a PDE and can perform inference orders of magnitude faster than conventional solvers \citep{lu2019deeponet,li2020fourier,brandstetter2022message, kovachki2023neural}. Most emulators are trained for a single PDE with fixed parameters, requiring a complete and costly retraining process for each new physical scenario.
However, creating effective surrogate models, a common form of empirical software in science, often involves a slow, manual process that bottlenecks scientific discovery. Recent AI systems aim to accelerate this by automatically generating and optimizing expert-level scientific software for scorable tasks using Large Language Models (LLMs) and Tree Search \cite{aygun2025ai}. Our work, as part of this broader effort, demonstrates this potential for AI-driven acceleration by focusing on automating the development of a generalized PDE emulator. 
Specifically, we used the system described in \cite{aygun2025ai} to develop a single, \textbf{generalized, equation-aware emulator}, a model that can accurately and stably simulate diverse physical systems, by encoding the target PDE in the model input. We condition the network's forward pass not only on the current state of the system but also on a compact vector representation -- an \textbf{equation encoding} -- that describes the structure of the PDE and the specific values of its parameters.

We report on four distinct, high-performing implementations generated by the automated AI system within this framework, spanning diverse approaches including architecture design, training paradigms, and hybrid modeling. These were evaluated on a challenging family of 1D PDEs from APEBench \citep{koehler2024apebench}: Advection-Diffusion, Burgers, Korteweg-de Vries (KdV), conserved Kuramoto-Sivashinsky (cKS), and Fisher's equation. We provide insights into their relative strengths, weaknesses, and generalization capabilities, particularly on a completely held-out PDE.

Our contributions are:
\begin{itemize}
    \item A unified framework for equation-aware emulation, studied with four modeling approaches.
    \item A demonstration of viability of learning a common mapping from PDE parameters to solution dynamics, by training each model on four distinct PDEs simultaneously.
    \item Generalization to a held-out PDE (Burgers' equation) while achieving strong performance on out-of-distribution parameter values, over many more rollout steps than in training.
    \item A explicit demonstration of how  an automated AI systems can accelerate the discovery and development of generalizable scientific models, achieving strong zero-shot performance in PDE emulation.
\end{itemize}

\subsection*{Related Work}

We address the generalization challenge faced by other work in this field. Approaches like \cite{gupta2022towards, takamoto2023learning} also explore parameter conditioning, but condition the model for a specific PDE family based on its scalar parameters (e.g., viscosity, force terms). 
Our equation-encoding scheme is more general. While the Physics Informed Token Transformer (PITT) \cite{lorsung2024physics} also ingests equation structure, it tokenizes symbolic text to learn an analytical correction and showed strong results when co-trained on several PDEs. Our method differs by using a compact, fixed-basis vector encoding  and by testing zero-shot generalization to entirely held-out PDEs, a scenario PITT did not test.
"Foundation models" for physics use large-scale pre-training on diverse datasets \cite{subramanian2023towards} but require fine-tuning to a specific downstream task via weight updates. We avoid fine-tuning by training a single model that takes the governing equation as a run-time input, and accurately models unseen PDEs. 
Other frameworks don't require fine-tuning, but do require significant examples at inference time to learn the new PDEs structure. E.g., the Multiple Physics Pretraining (MPP) framework of \cite{mccabe2023multiple} successfully trains a single model for predicting across multiple physical systems, but inference requires a history of prior states before it can infer the system's dynamics. Similarly, \cite{yang2023incontext} presents an in-context learning approach, which learns a new operator from example input-output pairs at inference time. Our "equation-prompted" approach allows us to model PDEs without any expensive context examples, yielding a system with good zero-shot generalization.
A related challenge is the high cost of training data generation. Active learning (AL) frameworks, like the AL4PDE benchmark \cite{musekamp2024active}, aim to improve data efficiency by intelligently selecting the most informative parameters and initial conditions for training. While this strategy optimizes training for a \textit{known} set of target PDEs, our work remains focused on inference-time generalization to \textit{novel, unseen} systems.

\section{Methodology}
Our methods operate under a unified framework. The task is to predict the system state $u(x, t+\Delta t)$ given the current state $u(x, t)$ and an equation encoding vector $\boldsymbol{c}$. The one-step prediction is:
$ \hat{u}(t+\Delta t) = \mathcal{M}(u(t), \boldsymbol{c}) $.
To encode the physics, we represent each PDE as a 7-dimensional vector encoding, $\boldsymbol{c} \in \mathbb{R}^7$, formed by the coefficients of a basis of common physical terms. Given a general 1D PDE of the form $u_t = F(u, u_x, u_{xx}, \dots)$, we define the basis using seven terms and their physical interpretations: linear reaction ($u$), non-linear reaction ($u^2$), linear advection ($u_x$), non-linear advection ($u u_x$), diffusion/viscosity ($u_{xx}$), dispersion ($u_{xxx}$), and hyper-diffusion ($u_{xxxx}$). E.g., Burgers' equation, $u_t = b u u_x + \nu u_{xx}$, is encoded as $\boldsymbol{c} = [0, 0, 0, b, \nu, 0, 0]$.

We study four distinct modeling techniques, each using our equation-conditioning mechanism but differing in architecture or training paradigm. See Appendix \ref{sec:appendix_models} for details.

\begin{itemize}[leftmargin=*]
    \item \textbf{PI-FNO-UNET:} A physics-informed U-Net \cite{ronneberger2015u} that operates on a multi-channel input composed of the state $u(t)$ and its pre-computed finite-difference derivatives. FNO blocks are conditioned via Feature-wise Linear Modulation (FiLM) \citep{perez2018film} and dynamically generated spectral weights.

    \item \textbf{LSC-FNO:} A latent operator \cite{wang2024latent} that learns features via a convolutional encoder-decoder. Core Fourier operator blocks are conditioned through: spectral gating, global attention, and FiLM \citep{perez2018film}.

    \item \textbf{PINO:} A training framework incorporating a physics-informed regularizer, inspired by \citep{li2021physics}. A scheduled PDE residual loss, calculated using spectral derivatives, is added to the primary objective to encourage physical consistency.

    \item \textbf{Learned Correction (LC):} A hybrid strategy where a FiLMed residual network \cite{he2016deep} learns to predict the correction for a fast but inaccurate coarse numerical solver. This approach reframes the task to learning the unresolved dynamics, which has proven highly data-efficient \citep{kochkov2021machine}.
\end{itemize}

\section{Experiments and Results}
We use the APEBench framework \cite{koehler2024apebench} to procedurally generate training and testing data. APEBench provides high-fidelity, differentiable solvers for a variety of PDEs.
We trained each of the models on a combined dataset generated from four 1D PDEs: Korteweg-de Vries (KdV), Conserved Kuramoto-Sivashinsky (cKS), Fisher's, and Advection-Diffusion equations, with different parameter sets (see Appendix for details). 
The models were evaluated on both in-distribution (ID) and out-of-distribution (OOD) parameters from these PDEs and on the entirely unseen Burgers' equation. 

\begin{figure}[h]
  \centering
  \includegraphics[width=\textwidth]{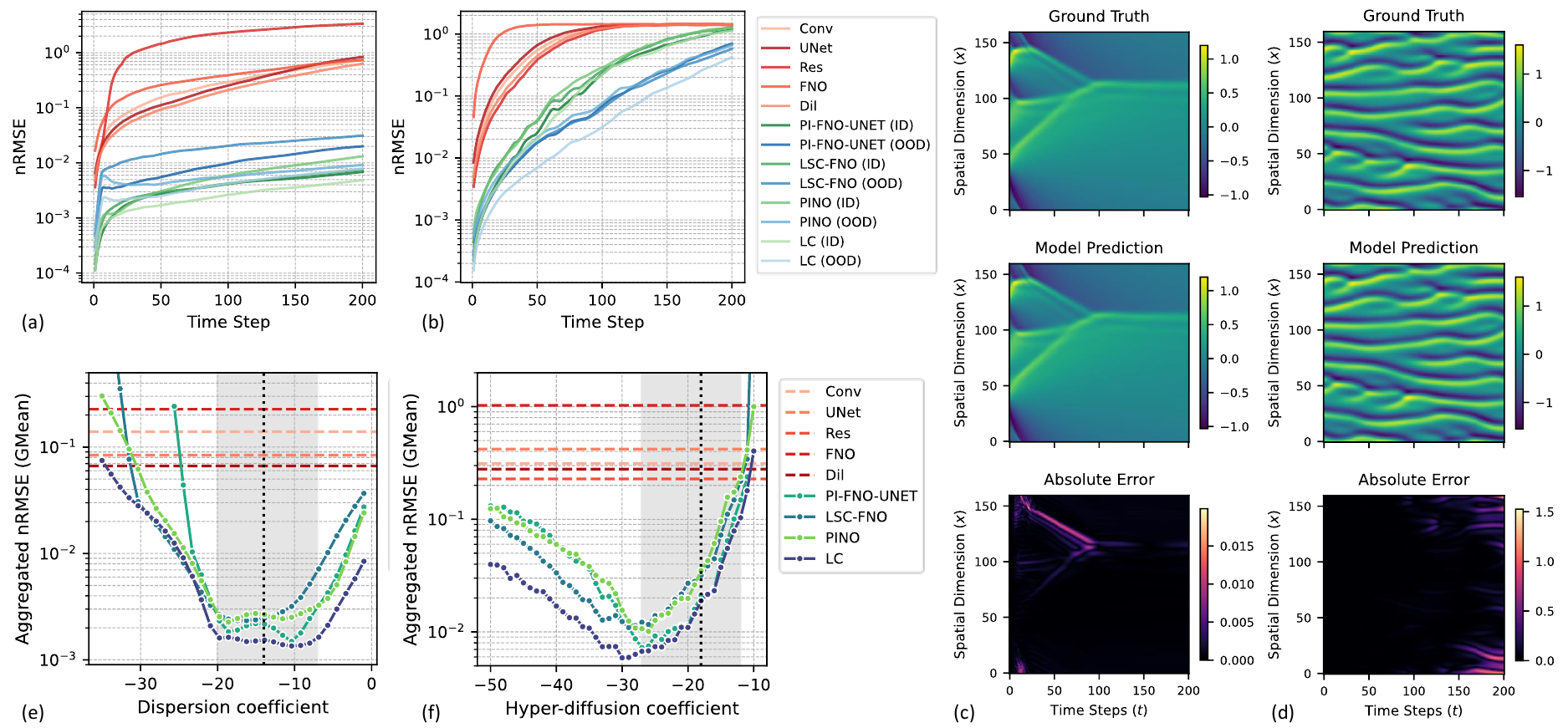}
  \caption{\textbf{Autoregressive rollout performance and OOD generalization. }
  Mean normalized RMSE over 200 timesteps for KdV (a) and cKS (b), averaged over 30 random initializations. Red curves are the baselines from \cite{koehler2024apebench}.  The generalized models are evaluated on both the ID parameter (green curves) used for training and an OOD parameter via zero-shot inference (blue curves).
  Representative rollouts from PI-FNO-UNET, evaluated on the OOD parameter for KdV (c) and cKS (d).
  Zero-shot generalization error versus the dispersion coefficient for KdV (e) and hyper-diffusion coefficient for cKS (f). The error is the aggregated nRMSE (Geometric Mean) over the first 100 steps. The shaded area represents the parameter range used to train the generalized models; the vertical dotted line indicates the parameter used to train the baselines \cite{koehler2024apebench}.}
  \label{fig:results}
\end{figure}

\subsection{Generalization to Unseen Parameters}
A generalized emulator should accurately predict system dynamics for parameters not seen during training. We show results for the Korteweg-de Vries (KdV) and Kuramoto-Sivashinsky (cKS) equations, which respectively test the model's ability to preserve stable soliton structures and predict complex chaotic dynamics. 

Figure \ref{fig:results} shows the rollout performance of the four models on both ID and OOD parameter sets. We compare our four generalized models against specialized baselines established in APEBench \cite{koehler2024apebench}. It is important to note that this is not intended as a direct, head-to-head performance benchmark; our models are trained on a diverse corpus spanning multiple PDEs, whereas the baselines are individually optimized for a single physical system. The baseline models are significantly smaller. The comparison instead demonstrates a more significant finding: our single, unified model, when applied in a zero-shot capacity without any task-specific fine-tuning, achieves competitive rollout performance with these expensive, specialized experts on both ID and OOD parameter sets. Our work establishes a strong baseline for multi-equation generalization, demonstrating that a single model can learn a transferable and robust understanding of physical dynamics.

Results are presented in Figure \ref{fig:results}. Panel (a,b) plot the normalized RMSE over a 200-step autoregressive rollout for KdV and cKS. By extending well beyond the 50-step rollouts used during training, we demonstrate strong temporal stability.
Panel (c,d) provides compares a ground truth rollout to a prediction from our PI-FNO-UNET model for an OOD parameter set of the KdV and cKS equations. The visualization confirms that the model accurately captures the defining physical behaviors for these unseen parameters: for KdV, it successfully models the complex soliton formation, merging, and interaction processes, while for KS, it reproduces the fine-grained chaotic spatial details with high fidelity.
Panel (e,f) quantifies the zero-shot generalization to OOD parameters, plotting the final rollout error as a function of KdV's dispersion coefficient and KS's hyper-diffusion coefficient. The training region is shaded, with the vertical dotted line indicating the single parameter value used to train the specialized baseline model. Our generalized models maintain low error far outside the training distribution, demonstrating a robust generalization capability that significantly surpasses the narrowly-tuned baseline.

A comparative analysis of our four generalized models reveals a performance hierarchy in the distinct strategies. The Learned Correction (LC) model consistently achieves the lowest error, consistent with \cite{kochkov2021machine}.
Its effectiveness stems from reframing the learning problem: by leveraging the coarse solver as a strong physical baseline, the network's task is simplified to predicting small, well-conditioned residual.
The remaining end-to-end emulators show comparable, excellent performance. The PINO model's success stems from its training objective, where a PDE residual loss regularizes the solution to adhere to the underlying physical laws. The LSC-FNO achieves its strong performance architecturally, using sophisticated latent-space conditioning with attention and spectral gating to automatically learn salient features for a given PDE. The PI-FNO-UNET, while still vastly outperforming all baselines, shows slightly higher error, suggesting that providing explicit derivative features as inputs, while a powerful inductive bias, may be a more rigid approach than the physics-regularized or adaptive feature-learning strategies.

\begin{figure}[h]
  \centering
  \includegraphics[width=\textwidth]{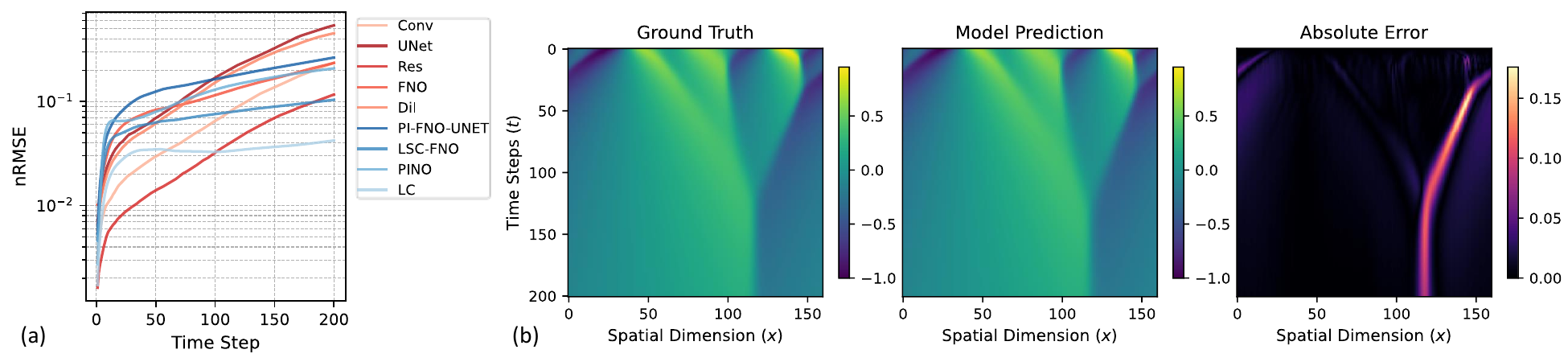} 
  \caption{
  \textbf{Zero-shot generalization on Burgers' equation.} 
  (a) nRMSE over 200 autoregressive time steps, averaged over 30 random initial conditions. Red curves are the baselines \cite{koehler2024apebench}. Blue curves are our generalized models.
  (b) A representative autoregressive rollout from the LSC-FNO model.
  }
  \label{fig:burgers_rollout}
\end{figure}

\subsection{Generalization to an Unseen PDE}

The key test of our framework is its ability to generalize not just to unseen parameters, but to an entirely new PDE. We held out the Burgers' equation, $u_t = b u u_x + \nu u_{xx}$, from the entire training corpus. Burgers' is a canonical model for shock formation, combining non-linear advection ($uu_x$) and diffusion ($u_{xx}$) operators that the models have encountered in training (e.g., in KdV and cKS) but never in combination. Notably, the coefficient for the second order derivative for cKS is negative(injection of energy), drastically different from the Burgers' equation evaluated here.

As shown in Figure \ref{fig:burgers_rollout}, all four of our generalized models produce stable and accurate rollouts for the Burgers' equation in a zero-shot setting. Their performance stands in stark contrast to the baseline models, which exhibit rapid error growth and instability.

This successful generalization provides strong evidence that our equation-aware framework learns more than just a mapping for specific PDE families, and suggests the models have learned a modular, compositional understanding of the underlying physics. The encoding for the Burgers' equation is prompting the model to compose its learned knowledge of the non-linear advection operator with its learned knowledge of the diffusion operator, learning a mapping from the coefficients of differential operators to their effect on system dynamics, a key step towards universal physical emulators.

\section{Automated Discovery via Tree Search}
The modeling approaches presented in this work were developed and refined using an AI system designed to automate the creation of expert-level empirical software for scorable scientific tasks \cite{aygun2025ai}. This system employs a LLM guided by a Tree Search (TS) algorithm to iteratively rewrite and improve code based on a defined quality metric. 
In our case, the scorable task was PDE emulation, and the quality metric was minimizing the nRMSE on a validation dataset comprising seen equations with in-distribution parameters.
This automated process allowed for the exploration of a vast space of potential architectures and training strategies. The system rapidly identified promising approaches, such as combinations of convolutional, residual, and spectral layers similar to FNO and U-Net, and organically discovered specific techniques like Feature-wise Linear Modulation (FiLM) for parameter conditioning when optimizing for generalization. The four distinct methods reported (PI-FNO-UNET, LSC-FNO, PINO, LC) are high-performing candidates discovered or refined through this automated search process.

\section{Conclusion}
We have presented a unified framework for building generalized, equation-aware PDE emulators. Our work, developed with the assistance of an AI system designed to automate scientific software creation \cite{aygun2025ai}, demonstrates that a single model trained on diverse physics can achieve remarkably competitive zero-shot performance against specialized experts. Rather than a direct performance benchmark, this work establishes a new, more challenging baseline for multi-equation generalization, highlighting the potential for AI-driven automation to accelerate the scientific discovery. This capability is underscored by the model's successful generalization to the unseen Burgers' equation, suggesting that our explicit equation encoding enables a learned, compositional understanding of differential operators.

Our systematic comparison of four strategies reveals a spectrum of powerful and practical techniques for building such universal emulators, spanning distinct architectural philosophies, training regularizers, and hybrid methods. These include a PI-FNO-UNET that directly ingests explicit physical derivative features as input, an advanced LSC-FNO that conditions a latent operator via attention and spectral gating, a PINO framework that regularizes training with a PDE residual loss, and a Learned Correction (LC) approach that augents coarse numerical solvers.
While promising for 1D PDEs, extending this framework to higher-dimensional systems is necessary but faces scaling challenges due to increased complexity and computational cost. Our exploration, especially into higher dimensions, was partly constrained by the computational resources available in the AI agent's development sandbox environment. Future work should also assess applicability to broader PDE classes and boundary conditions. Despite these limitations, this work demonstrates a viable path towards universal PDE emulators informed directly by governing equations, paving the way for more efficient scientific simulation tools.

\begin{ack}
This work utilized the APEBench framework for data generation and benchmarking. We thank the developers for making this valuable tool publicly available. We thank Dmitrii Kochkov, Ryan Krueger, Francesco Mottes, for helpful discussion.

\end{ack}

\bibliographystyle{plainnat}
\bibliography{main}


\appendix

\section{Experimental Setup}

\subsection{Data generation}
All data were generated on-the-fly using APEBench \cite{koehler2024apebench}. 
The model was trained on data generated from the four PDEs listed below, with parameters sampled across a grid to create a diverse dataset. 
\begin{enumerate}
    \item \textbf{Korteweg-de Vries (KdV):} $u_t = b u u_x + \epsilon u_{xxx} + \zeta u_{xxxx}$
    \item \textbf{Conserved Kuramoto-Sivashinsky (cKS):} $u_t = b u u_x + \nu u_{xx} + \zeta u_{xxxx}$ ($\nu<0$ causing instability)
    \item \textbf{Fisher's Equation:} $u_t = \nu u_{xx} + r u(1-u)$
    \item \textbf{Advection-Diffusion:} $u_t = c u_x + \nu u_{xx}$
\end{enumerate}
To evaluate generalization, Burgers' equation was held out as an entirely unseen test case. 
\begin{itemize}
    \item \textbf{Burgers' Equation (Held-Out):} $u_t = b u u_x + \nu u_{xx}$
\end{itemize}
The parameter ranges for each simulation, guided by the difficulty parameters in APEBench, are detailed below:
\begin{itemize}
    \item \textbf{KdV:} $b \in [-2.0, -1.0]$, $\epsilon \in [-20.0, -7.0]$, $\zeta \in [-9.0, -3.0]$.
    \item \textbf{cKS:} $b \in [-2.0, -1.0]$, $\nu \in [-2.0, -0.5]$, $\zeta \in [-27.0, -12.0]$.
    \item \textbf{Fisher:} $r \in [0.01, 0.05]$, $\nu \in [0.2, 5.0]$ (with the coefficient of $u^2$ being $-r$).
    \item \textbf{Advection-Diffusion:} $c \in [-4.0, 4.0]$, $\nu \in [2.0, 8.0]$.
\end{itemize}
For each set of parameters, we randomly generate 50 samples with 50 steps with different initial conditions. The test performance is evaluated on 30 samples with random initial conditions, rolled out for 200 steps.
We use the default setting in \cite{koehler2024apebench} -- all simulations were performed on a spatial domain of size $L=1.0$ discretized with 160 points, and a time step of $\Delta t=1$. For the PINO model, training was performed on unrolled sequences of 5 time steps.
This split was chosen to test compositional generalization, as the held-out Burgers' equation is a unique combination of operators from the training set. It shares its non-linear advection term ($u u_x$)with the KdV and cKS equations, and its stabilizing, positive diffusion term ($\nu>0$) with the Fisher's and Advection-Diffusion equations. 
This presents a non-trivial generalization challenge, particularly concerning the shared terms with cKS. 
The sign of the diffusion coefficient $\nu$ is crucial, as it fundamentally reverses the dynamics. In Burgers equation, $\nu > 0$ provides stabilizing diffusion (viscosity). Conversely, in cKS, the same operator form with $\nu < 0$ acts as anti-diffusion, actively driving instability.
This challenges the model to synthesize a new system by correctly applying the principle of stabilizing diffusion, rather than the instability-driving dynamics of the similar-looking term in cKS.

\subsection{Evaluation Metric}
The primary evaluation metric is the normalized Root Mean Squared Error (nRMSE), defined as:
$$ \text{nRMSE}(t) = \frac{\| u_{\text{pred}}(t) - u_{\text{true}}(t) \|_2}{\| u_{\text{true}}(t) \|_2} $$
averaged over all samples.
For the aggregated errors shown in Fig.\ref{fig:results}(e,f), we use the Geometric Mean of the Normalized Root Mean Squared Error (GMean of nRMSE), aggregated over 100 steps of the test rollout. It is defined as $$\text{Aggregated nRMSE} = \exp\left(\frac{1}{T} \sum_{t=1}^{T} \log(\text{nRMSE}(t))\right)$$

\section{Automated Discovery via Tree Search}
\label{sec:appendix_treesearch}
The TS process began with an initial code implementation, often based on a basic template, or sometimes a standard baseline architecture (like FNO, U-Net, ResNet). The system then iteratively performed the following steps:
\begin{enumerate}
    \item \textbf{Node Selection:} The TS algorithm selected a promising code candidate (a node in the tree) based on a balance of exploitation (high past scores) and exploration (visiting less explored code variations), using a PUCT strategy \cite{silver2017mastering}.
    \item \textbf{Code Rewriting:} The selected code, along with the task description, evaluation metrics, and potentially specific instructions or "research ideas" (e.g., implement FiLM conditioning, use unrolled training, focus on equation encoding), was fed to an LLM. The LLM generated a modified version of the code.
    \item \textbf{Evaluation:} The new code was executed in a sandboxed environment, trained on the training dataset, and evaluated on the validation dataset to obtain a quality score (e.g., aggregated nRMSE across validation scenarios). Note that the validation dataset contains only seen equations with in-distribution parameters.
    \item \textbf{Tree Update:} The new code, its score, and execution logs were added as a new node to the tree, and the visit counts and scores were updated back up the tree.
\end{enumerate}
More details can be found in \cite{aygun2025ai}.

\section{Model and Training Details}
\label{sec:appendix_models}
The source code, datasets, and scripts for reproducing the figures are available in \href{https://github.com/google-research/generalized-pde-emulator}{\texttt{https://github.com/google-research/generalized-pde-emulator}}.
We used an AI agent to help implement and advance the methods as described in Appendix \ref{sec:appendix_treesearch}.

Key hyperparameters are summarized in Table \ref{tab:hyperparams}. 
Below we provide the more detailed architectural descriptions.

\begin{table}[h!]
  \caption{Comparative Model and Training Hyperparameters}
  \label{tab:hyperparams}
  \centering
  \begin{tabularx}{\textwidth}{lXXXX}
    \toprule
    \textbf{Parameter} & \textbf{PI-FNO-UNET (M1)} & \textbf{LSC-FNO (M2)} & \textbf{PINO (M3)} & \textbf{LC (M4)} \\
    \midrule
    \multicolumn{5}{l}{\textit{\textbf{Architecture}}} \\
    Core Operator & FiLMed FNO U-Net & Gated FNO (LNO) & FiLMed FNO & FiLMed ResNet \\
    Input Channels & 7 (u + derivs) & 1 (u) & 3 (u + encoding) & 49 (u + features) \\
    Latent Channels & 128 & 128 & 256 & 160 \\
    Depth / Blocks & 4 levels (U-Net) & 12 blocks & 6 blocks & 14 blocks \\
    Activation & SiLU & SiLU & SiLU & GeLU \\
    \midrule
    \multicolumn{5}{l}{\textit{\textbf{Training}}} \\
    Training Steps & 100,000 & 100,000 & 100,000 & 100,000 \\
    Batch Size & 64 & 128 & 64 & 128 \\
    Peak Learning Rate & $4 \times 10^{-4}$ & $4 \times 10^{-4}$ & $3 \times 10^{-4}$ & $5 \times 10^{-4}$ \\
    Loss Function & MAE & MAE & MAE + PINO & MAE \\
    PINO Max Weight & N/A & N/A & $3 \times 10^{-3}$ & N/A \\
    \bottomrule
  \end{tabularx}
\end{table}

\paragraph{Method 1: Physics-Informed U-Net with FNO layer (PI-FNO-UNet)}
This approach directly incorporates physical knowledge by pre-computing derivatives of the input state $u(t)$ using finite differences. The input to the network is a 7-channel tensor formed by concatenating the state $u$ with its numerically computed non-linear terms ($u^2, u u_x$) and spatial derivatives ($u_x, u_{xx}, u_{xxx}, u_{xxxx}$). The network itself is a U-Net architecture composed of \textbf{FiLMed-FNO Blocks}. The equation encoding $\boldsymbol{c}$ conditions the network in two ways:
\begin{enumerate}[leftmargin=*]
    \item \textbf{Dynamic Spectral Weights:} The weights of the spectral convolution layers are dynamically generated by an MLP that takes the encoding $\boldsymbol{c}$ as input. This allows the model to learn a direct mapping from PDE parameters to the appropriate Fourier-space linear operator.
    \item \textbf{Feature-wise Linear Modulation (FiLM):} FiLM layers \citep{perez2018film} use $\boldsymbol{c}$ to apply a feature-wise affine transformation after spatial convolutions, providing powerful conditioning in the physical domain.
\end{enumerate}
The U-Net structure enables multi-scale feature processing, which is beneficial for capturing dynamics across different spatial frequencies.

\paragraph{Method 2: Latent Operator with Attention and Gating (LSC-FNO)}
Instead of relying on hand-engineered features, this architecture learns relevant spatial features automatically. A strided convolutional encoder projects the input state $u(t)$ into a lower-resolution, higher-channel latent space. The core of the model is a sequence of operator blocks that process this latent representation, conditioned by the encoding $\boldsymbol{c}$ via three distinct mechanisms:
\begin{enumerate}[leftmargin=*]
    \item \textbf{Spectral Gating:} The Fourier modes of the latent state are multiplied by a \enquote{gate} vector produced from the encoding $\boldsymbol{c}$. This allows the model to selectively filter or amplify frequencies based on the PDE's characteristics.
    \item \textbf{Global Attention:} We use a multi-head attention mechanism where the latent state serves as the query, and the encoding $\boldsymbol{c}$ is projected to form the key and value. This allows the model to globally modulate all spatial features based on the governing equation.
    \item \textbf{Feature-wise Linear Modulation (FiLM):} As in Method 1, FiLM layers provide direct conditioning of the latent features.
\end{enumerate}
A corresponding transposed convolutional decoder projects the final latent state back to the original spatial resolution to produce the prediction.
\paragraph{Method 3: Physics-Informed Regularization (PINO)}
Long-term stability is a major challenge for autoregressive PDE emulators. To mitigate error accumulation, we incorporate a physics-informed loss term, inspired by Physics-Informed Neural Operators (PINO) \citep{li2021physics}. The total loss function combines a standard data-driven loss with a penalty for violating the governing PDE:
$$ \mathcal{L}_{\text{total}} = \mathcal{L}_{\text{data}} + \lambda(s) \cdot \mathcal{L}_{\text{PDE}} $$
where $s$ is the training step. $\mathcal{L}_{\text{data}}$ is the mean absolute error (MAE) between the predicted and true future states, computed over a short rollout of 5 steps. $\mathcal{L}_{\text{PDE}}$ is the MAE of the PDE residual, calculated by applying spectral differentiation to the \textit{ground truth} data and comparing the time derivative with the right-hand side of the equation defined by the encoding $\boldsymbol{c}$. The weight $\lambda(s)$ is scheduled to increase from 0 to a maximum value during training, gradually enforcing physical consistency.
The underlying architecture is a FiLMed FNO. The input to the network is a multi-channel tensor that combines the normalized state $u(t)$ with a learned linear projection of the equation encoding. 

\paragraph{Method 4: Hybrid Emulation via Learned Correction (LC)}
This strategy reframes the problem from direct emulation to learning the error of a simpler, traditional model, which has proven to be much more efficient than learning the solution from scratch \cite{kochkov2021machine}. For each PDE, APEBench provides a fast but less accurate \enquote{coarse stepper} (e.g., a numerical solver with a large time step). Our model is trained to predict the correction term required to match the high-fidelity solution.
The model, a deep FiLMed residual network with spectral convolutions in its residual blocks, takes the coarse prediction and the equation encoding $\boldsymbol{c}$ as inputs and outputs the predicted correction. This hybrid approach leverages the efficiency of the coarse solver while using the neural network to learn the complex, unresolved dynamics. The final prediction is the sum of the coarse prediction and the network's output.

\end{document}